\documentclass[twocolumn,twoside,a4paper, 10pt]{IEEEtran}

\usepackage{subcaption}
\usepackage{multirow}

\usepackage{multirow}
\usepackage{graphicx}
\usepackage[margin=1in]{geometry}
\usepackage{array}
\usepackage{grffile}

\title{Enhancing Underwater Images Using Deep Learning with Subjective Image Quality Integration}

\author{Jose M. Montero, Jose-Luis Lisani \\
{\small Universitat de les Illes Balears, IAC3, Spain (joseluis.lisani@uib.es)}}

\usepackage[font={sf,footnotesize}]{caption}

\usepackage{hyperref,graphicx,amsmath,amssymb,amsthm}
\usepackage{url}

\usepackage{titlecaps}
\Addlcwords{a an on and for the of in to with from which as his their at or those by its et ses des de du la en les aux le dans zu l0}

\graphicspath{{./figures/}}

\begin{document}

\maketitle

\begin{abstract}
Recent advances in deep learning, particularly neural networks, have significantly impacted a wide range of fields, including the automatic enhancement of underwater images. This paper presents a deep learning-based approach to improving underwater image quality by integrating human subjective assessments into the training process. To this end, we utilize publicly available datasets containing underwater images labeled by experts as either high or low quality. Our method involves first training a classifier network to distinguish between high- and low-quality images. Subsequently, generative adversarial networks (GANs) are trained using various enhancement criteria to refine the low-quality images. The performance of the GAN models is evaluated using quantitative metrics such as PSNR, SSIM, and UIQM, as well as through qualitative analysis. Results demonstrate that the proposed model—particularly when incorporating criteria such as color fidelity and image sharpness—achieves substantial improvements in both perceived and measured image quality.
\end{abstract}

\section{Introducción}
\label{sec:intro}
In recent years, the rapid advancement of deep learning—particularly through neural networks—has revolutionized numerous fields. At the core of these models lies data: the quality and quantity of available training samples directly influence a network’s ability to produce accurate and meaningful outcomes. In the domain of underwater image enhancement, this requires access to low-quality underwater images along with corresponding high-quality references to guide the training process.

However, acquiring such paired data under real underwater conditions is particularly challenging. Early efforts circumvented this issue by simulating underwater distortions: high-quality in-air images were artificially degraded using underwater image formation models to produce synthetic training pairs~\cite{P2P, uwcnn, anwar2018deep}. Unfortunately, models trained on synthetic datasets often failed to generalize to real-world underwater scenes, sometimes performing worse than basic classical enhancement techniques~\cite{anwar2020diving}.

To address this, several datasets have been introduced that better reflect real underwater environments. Notably, the UIEB dataset~\cite{UIEB} includes real underwater images labeled as low quality, along with corresponding enhanced versions produced using various classical algorithms. For each image, the best enhancement was selected by expert consensus to represent the most visually acceptable result. The EUVP dataset~\cite{EUVP} expanded on this by including both unpaired (high and low quality, selected via expert agreement) and paired images, where high-quality references were degraded using specific models to generate low-quality counterparts. Consequently, both datasets embed elements of subjectivity, reflecting human judgments on visual quality.

This work investigates deep neural network architectures for the analysis and enhancement of underwater images by explicitly incorporating human subjectivity into the learning process. First, we propose a classifier trained to distinguish between high- and low-quality underwater images, emulating the subjective labeling present in UIEB and EUVP.

\begin{figure}[h]
\begin{center}
	\includegraphics[width = \linewidth]{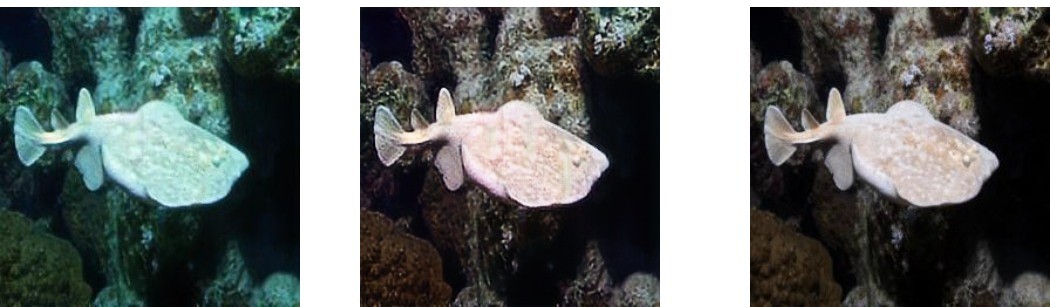}
\end{center}
\caption{Example of an underwater image (left) and the result of our enhancement model (center). The corresponding high-quality version of the input is also shown (right). This image was not used during training, but a result similar to the target was obtained.} \label{fig:fig1}
\end{figure}

Second, we integrate this classifier into an enhancement model, trained to improve low-quality underwater images. Two approaches are considered: one trained on unpaired data, guided solely by the classifier to reflect subjective preferences; and another trained on paired data, where high-quality images serve as explicit enhancement targets. Figure~\ref{fig:fig1} illustrates a representative output from the final enhancement model.

The paper is organized as follows: Section~\ref{sec:relatedwork} reviews existing underwater image enhancement techniques using deep learning. Sections~\ref{sec:classifier}, \ref{sec:gan}, and~\ref{sec:tests} detail the proposed classification and enhancement models, along with experimental results. Finally, conclusions are presented in Section~\ref{sec:conclusions}.

\section{Related Work}
\label{sec:relatedwork}

This research falls within the domain of automatic underwater image enhancement techniques powered by deep learning. These methods aim to improve the clarity, perceptual quality, and visual detail of underwater imagery, which is critical for applications such as marine research, underwater robotics, and maritime surveillance.

A major body of work closely aligned with our approach involves the use of convolutional neural networks (CNNs) to address the unique challenges of underwater image degradation. These methods typically employ encoder-decoder architectures that learn to suppress noise and enhance visual detail while preserving essential structural features.

An extensive review of deep learning-based underwater enhancement techniques was presented in~\cite{anwar2020diving}, which emphasized the field's heavy reliance on synthetic training data due to the scarcity of real-world paired datasets~\cite{P2P, uwcnn, WaterGan}. In contrast, our work leverages real underwater images and introduces new optimization criteria to improve generalization and performance in real-world scenes.

Among the models surveyed in previous work, UGAN~\cite{Ugan} is particularly relevant. UGAN utilizes a generative adversarial network (GAN) framework, where a generator attempts to produce realistic enhanced images capable of ``fooling'' a discriminator. The discriminator in UGAN adopts the WGAN-GP architecture~\cite{WGAN}, which incorporates a gradient penalty to promote stable and realistic outputs. It is fully convolutional and follows a PatchGAN~\cite{patch} design, evaluating local image patches rather than entire images—thus improving sensitivity to fine-grained textures.

The UGAN generator is inspired by CycleGAN~\cite{Cycle} and resembles a U-Net architecture~\cite{Unet}. CycleGAN enables image-to-image translation without requiring paired examples, making it especially suitable for underwater enhancement where ground-truth images are often unavailable.

UGAN is trained and evaluated on subsets of ImageNet~\cite{ImNet}, using input images sized at $256\times256\times3$ pixels. The dataset consists of 6,143 undistorted and 1,817 distorted underwater images.

DenseGAN~\cite{DenseGan} is another relevant model, combining a generator and a discriminator similar in structure to PatchGAN. However, it introduces five spectral normalization layers~\cite{espectral} in the discriminator to stabilize training and better detect subtle visual artifacts. DenseGAN also uses ImageNet for training, with the same input dimensions.

FUnIE-GAN~\cite{EUVP}, introduced alongside the EUVP dataset, proposes a GAN-based framework with a U-Net-inspired generator that incorporates encoder-decoder skip connections to preserve image details. The discriminator follows a Markovian PatchGAN design~\cite{isola2018imagetoimage}, which assesses local realism while preserving spatial continuity across adjacent patches. FUnIE-GAN is trained on 11,000 paired and 7,500 unpaired underwater images.

In parallel, Water-Net~\cite{UIEB}, trained on the UIEB dataset, uses a convolutional fusion network that learns three confidence maps to combine enhanced versions of an input image—produced using white balancing, histogram equalization, and gamma correction—into a final output. This fusion mechanism allows the network to adaptively emphasize different visual properties. Water-Net is trained on 800 paired images using a content loss derived from VGG19 features~\cite{netVgg}, encouraging structural and perceptual similarity between enhanced and reference images.

Our approach draws inspiration from UGAN~\cite{Ugan} and Water-Net~\cite{UIEB}, with several key distinctions. Initially, we trained a classifier to distinguish between high- and low-quality underwater images, mimicking the expert labeling process used in the UIEB and EUVP datasets. This classifier is described in detail in the next section.

Subsequently, the classifier was integrated with a generator in a GAN-like framework, although only the generator was updated during training. The goal was to refine images until they were classified as high quality by the discriminator. However, this strategy achieved limited success. As a solution, we incorporated paired data to explicitly guide the training process via supervised loss. Finally, we adopted a conventional adversarial setup, updating both generator and discriminator jointly. The architectures and results are presented in Section~\ref{sec:tests}.

\section{Classifier}
\label{sec:classifier}

The objective of the classifier is to emulate human subjectivity in distinguishing between high- and low-quality underwater images. To this end, we train the model exclusively on real, unpaired images previously labeled by expert consensus. Specifically, we use the unpaired subset of the EUVP dataset~\cite{EUVP}, as these images have not been preprocessed or artificially enhanced. This selection enables the classifier to learn and internalize the nuanced, subjective criteria that experts rely on when evaluating underwater image quality.

\subsection{Model}

Our goal is to learn a function $D: X \rightarrow \text{Score}$, where $X$ denotes the space of underwater images with varying levels of visual quality. The function outputs a continuous score indicating perceived image quality: values closer to 0 correspond to high quality, while values closer to 1 indicate low quality. A threshold of 0.5 is used to classify images into high- or low-quality categories.

\begin{figure*}
\begin{center}
	\includegraphics[width = \linewidth]{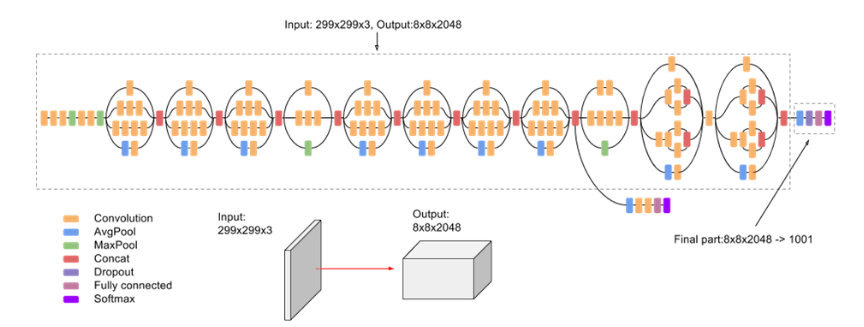}
\end{center}
\caption{Network architecture of InceptionV3~\cite{InceptionV3}.}
\label{fig:Discriminador}
\end{figure*}

For this task, we selected the InceptionV3 network~\cite{InceptionV3}, a deep convolutional neural network originally designed for large-scale image classification tasks such as those in the ImageNet dataset~\cite{ImNet}. The architecture of InceptionV3, illustrated in Figure~\ref{fig:Discriminador}, comprises a sequence of convolutional and pooling layers with varying filter sizes, enabling it to extract multi-scale features and capture both local and global structures in input images.

InceptionV3 was chosen for its strong feature representation capabilities and its demonstrated generalization across diverse image domains. It achieved 78.1\% top-1 accuracy on the ImageNet benchmark, which includes over one million labeled images across a wide range of categories. This generalization capability is particularly relevant to our application, where subjective visual characteristics are central to perceived image quality.

To adapt InceptionV3 for our binary classification task, we modified its output layer to distinguish only between high- and low-quality underwater images. This streamlined configuration allows the network to function as an effective surrogate for human evaluators, capturing subtle visual cues that correspond to expert judgments. Additionally, the classifier serves a dual purpose in our framework: it not only predicts image quality but also guides the training of generative models designed to enhance low-quality underwater imagery.

\subsection{Training}

As previously mentioned, we trained the classifier using the unpaired real-image subset from the EUVP dataset~\cite{EUVP}, which includes a total of 7,375 images—3,720 labeled as low quality and 3,655 as high quality. We randomly selected 515 images from each class for validation, leaving 3,205 low-quality and 3,140 high-quality images for training.

We applied data augmentation techniques, including horizontal and vertical flipping. However, we chose not to normalize the images, as color fidelity is a key indicator of image quality in underwater environments. Early stopping was used with a patience of 5 epochs, allowing the training to terminate if no improvement was observed on the validation set for five consecutive epochs. Training was conducted for a maximum of 40 epochs, with a learning rate of 0.001, using the ADAM optimizer~\cite{adam}—a well-established method for stable convergence. Binary cross-entropy~\cite{binacryCross}, commonly used for binary classification tasks, was employed as the loss function. The final model was obtained after 30 epochs, based on early stopping.

To evaluate classifier performance, we used the paired subset of the EUVP dataset, which was not involved in training. The following metrics were used for evaluation~\cite{arce-2022}:

\begin{equation}\label{FormulasMetricas}
\begin{aligned}
    \text{Precision} &= \frac{TP}{TP + FP} \\
    \text{Recall} &= \frac{TP}{TP + FN} \\
    \text{Specificity} &= \frac{TN}{TN + FP} \\
    \text{Accuracy} &= \frac{TP+TN}{TP+TN+FP+FN}
\end{aligned}
\end{equation}

In these formulas, TP (true positives) are high-quality images correctly classified as such, and FP (false positives) are low-quality images incorrectly classified as high quality. TN (true negatives) represent correctly classified low-quality images, while FN (false negatives) refer to high-quality images misclassified as low quality.

Accuracy measures the overall proportion of correct predictions and is particularly meaningful when class distributions are balanced. However, for reliable classification, both recall and specificity must also be high: high recall ensures most high-quality images are correctly identified, while high specificity ensures most low-quality images are accurately rejected. This balance reduces false positives and false negatives, improving overall precision and minimizing misclassification risk.

Validation was performed using the EUVP paired dataset, which consists of three subsets: `Underwater Scenes' (4,370 images; 2,185 per class), `Underwater ImageNet' (7,400 images; 3,700 per class), and `Underwater Dark' (11,100 images; 5,500 per class). Together, these subsets total 22,870 images. Additionally, the UIEB dataset was used for further validation, contributing 1,780 images (890 per class). All these images were included in the final evaluation. The results are presented in Table~\ref{tablaDiscriminador}.

\begin{table}[h!]
\centering
\renewcommand{\arraystretch}{1.5}
\begin{tabular}{|>{\raggedright}m{1.5cm}|c|c|c|c|}
\hline
\multirow{2}{*}{} & \multicolumn{2}{c|}{\textbf{EUPV}} & \multicolumn{2}{c|}{\textbf{UIEB}} \\
\cline{2-5}
& \shortstack{ \\ Good \\ Quality} & \shortstack{Bad \\ Quality} & \shortstack{Good \\ Quality} & \shortstack{Bad \\ Quality} \\
\hline
Precision & 0.95 & 0.93 & 0.81 & 0.83 \\
\hline
Recall / Specificity & 0.94 & 0.95 & 0.84 & 0.80 \\
\hline
Accuracy & \multicolumn{2}{c|}{0.94} & \multicolumn{2}{c|}{0.82} \\
\hline
\end{tabular}
\caption{Performance comparison between EUPV and UIEB datasets.}
\label{tablaDiscriminador}
\end{table}

As shown in Table~\ref{tablaDiscriminador}, the classifier trained on the unpaired EUVP dataset demonstrates strong performance, with all metrics exceeding 0.9 on the EUVP subsets and over 0.8 on the UIEB dataset. The classifier outputs a continuous quality score between 0 and 1, where values below 0.5 indicate high-quality images and values above 0.5 denote low quality, as described previously.

\begin{figure}[!htb]
	\begin{center}
		\includegraphics[width = 170px]{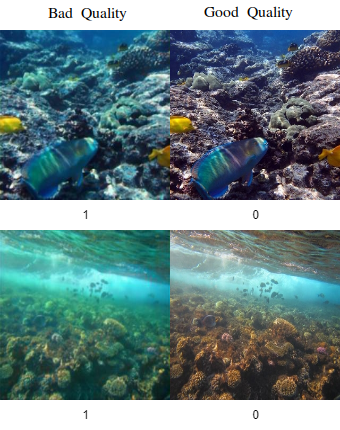}
	\end{center}
	\caption{Examples of underwater images scored by the classifier.}
	\label{resultadosClass}
\end{figure}

Figure~\ref{resultadosClass} shows two low-quality underwater images, each receiving a classifier score of 1 due to their lack of sharpness and poor color fidelity. Their corresponding reference images, also shown, received scores of 0, reflecting good visual quality characterized by appropriate sharpness and color correction. All examples belong to the paired EUVP validation dataset.

\section{GAN}
\label{sec:gan}

\begin{figure}[h]
\begin{center}
	\includegraphics[width = 240px]{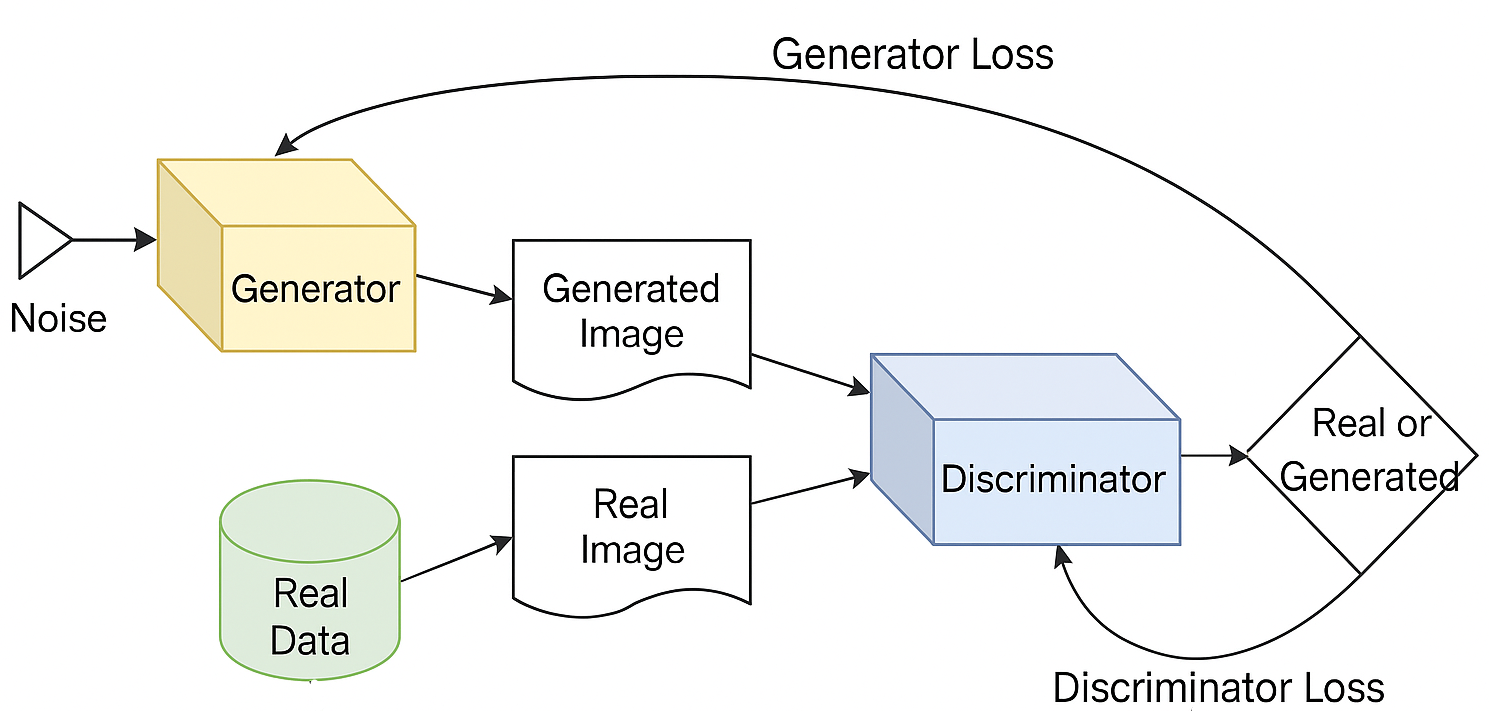}
\end{center}
\caption{Structure of a generative adversarial network.}
\label{GanARQUICTERURA}
\end{figure}

Generative Adversarial Networks (GANs) are machine learning models composed of two competing neural networks: a generator (G) and a discriminator (D), the latter typically implemented as a classifier. The generator—often designed as an encoder-decoder architecture~\cite{Unet}—produces synthetic data resembling real samples, while the discriminator evaluates the authenticity of the generated outputs. These networks are trained simultaneously in a competitive setting, gradually improving through adversarial feedback~\cite{GAN}. GANs have been successfully applied to tasks such as image generation, super-resolution, and domain translation~\cite{GANoVER}. The general structure of a GAN is illustrated in Figure~\ref{GanARQUICTERURA}.

\subsection{Model}

Our objective is to learn a mapping function $G: X \rightarrow Y$, where the input domain $X$ consists of low-quality underwater images, and the target domain $Y$ comprises their high-quality counterparts. The generator learns this mapping by iteratively improving its outputs to minimize both the adversarial loss from the discriminator and other reconstruction losses.

\begin{figure*}
\begin{center}
	\includegraphics[width = \linewidth]{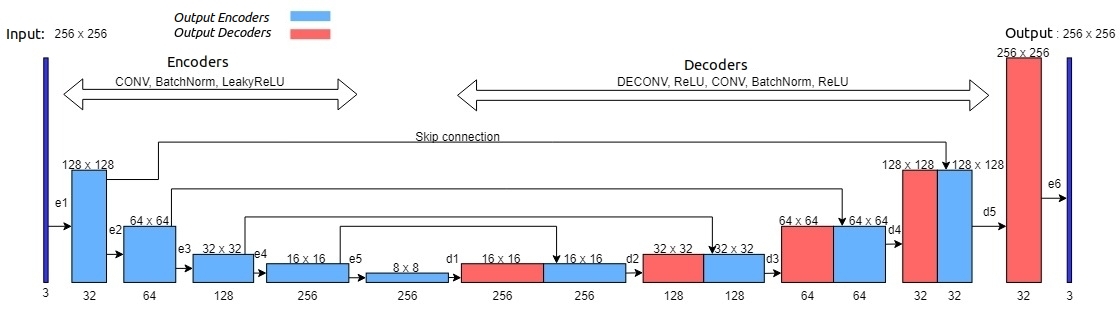}
\end{center}
\caption{Architecture of the proposed generator network.}
\label{fig:generator}
\end{figure*}

The architecture of our generator, shown in Figure~\ref{fig:generator}, is based on the U-Net model~\cite{Unet}, a widely used encoder-decoder architecture. It comprises a series of convolutional encoder layers (e1–e5) and decoder layers (d1–d5), with skip connections between encoder and decoder layers of matching resolution (e.g., (e1, d4), (e2, d3), etc.). These skip connections allow the decoder to access low-level spatial features, which improves image reconstruction and preserves important visual details. This architectural strategy has proven effective in domain translation tasks involving underwater images~\cite{laxman2021efficient, torbunov2022uvcgan, isola2018imagetoimage}.

In our case, we opted for a slightly smaller network due to the relatively low resolution of our input images ($256 \times 256 \times 3$). This design choice also improves computational efficiency, which is beneficial for deployment in real-time marine scenarios. The output of the encoder (e1–e5) produces a compact feature representation of size $8 \times 8 \times 256$, which is then passed through the decoder layers (d1–d6), along with the features provided via skip connections, to reconstruct an enhanced image of the same original size.

Each encoder layer applies 2D convolutions with a kernel size of $4 \times 4$, followed by batch normalization (BatchNorm2d) to stabilize training and improve generalization~\cite{_2022_qu}, and Leaky ReLU activation, which prevents gradient vanishing in negative regions~\cite{javi_2023_la}. The decoder layers use transposed convolutions (ConvTranspose2d) with a $4 \times 4$ kernel to upsample the image. These are followed by ReLU activations and additional convolutional layers (with $3 \times 3$ kernels) that preserve spatial resolution while allowing the network to learn more complex features. Batch normalization and ReLU are again applied to maintain stability and introduce non-linearity.

\subsection{Loss Function}

The adversarial loss function~\cite{GAN} associated to a generic GAN model is defined as:

\begin{equation} 
\begin{split}
L_{\text{GAN}}(G,D) = &\ \mathbb{E}_{x \sim p_{\text{data}}(x)} [\log D(x, y)] + \\
                      &\ \mathbb{E}_{x \sim p_x(x)} [\log(1 - D(G(x)))],
\end{split}
\end{equation}
where $p_{\text{data}}(x)$ denotes the distribution of real data, and $p_x(x)$ the distribution of generated data. The term $D(x, y)$ evaluates real images and their corresponding labels, while $D(G(x))$ evaluates synthetic images produced by the generator using a noise image as input. The generator $G$ attempts to minimize this loss by producing images that the discriminator $D$ classifies as real, while $D$ is trained to distinguish between real and generated samples.

In the context of this paper, the \textit{real} images are good-quality underwater images and the generator modifies low-quality inputs until the discriminator classifies them as good.

Notably, this training framework does not require paired image data; the generator can be trained solely with poor-quality inputs, adjusting its outputs until they are recognized as high-quality by the discriminator. 
However, since both EUVP and UIEB datasets provide paired examples of low- and high-quality underwater images, we further enhanced the training process by incorporating additional loss functions. These losses encourage improved similarity with reference images, better color correction, and sharper details. In all formulations below, $X$ denotes the low-quality input image and $Y$ the corresponding high-quality reference image.

\begin{itemize}
    \item \textbf{Global similarity}:  
    Prior work in image-to-image translation~\cite{9744097,isola2018imagetoimage} has shown that L1 and L2 losses effectively encourage the generator to map inputs from $X$ to outputs in $Y$.  L1 loss better preserves fine details by penalizing large deviations less aggressively.  L2 loss encourages smoother results, reducing high-frequency noise.

    We implemented and evaluated both losses separately. Their mathematical formulations are:
    \begin{align}
    L_1(X, Y) &= \mathbb{E}_{X,Y} [\|Y - G(X)\|_1] \\
    L_2(X, Y) &= \mathbb{E}_{X,Y} [\|Y - G(X)\|_2^2]
    \end{align}
    \item \textbf{Color correction (Angular loss)}:  
    Following the approach in~\cite{10.3389/fmars.2023.1161399}, we integrated angular loss to enhance the naturalness of color in generated images. This loss measures the angle between the RGB vectors of $G(X)$ and $Y$, reducing color distortion:

    \begin{equation}
    L_{\text{angular}}(X, Y) = \mathbb{E}_{X,Y} [\angle(G(X), Y)]
    \end{equation}

    \item \textbf{Sharpness (Gradient loss)}:  
    Generative models often produce slightly blurry images. To mitigate this, we employed a gradient difference loss (GDL), as in~\cite{mathieu2016deep}, which compares local spatial gradients in the target and generated images. With $\alpha = 1$, the formulation is:

    \begin{equation}
    \begin{split}
      &L_{gdl}(X, Y) = \\
     & \sum_{i,j} (|Y_{i,j} - Y_{i-1,j}| - 
     |G(X)_{i,j} - G(X)_{i-1,j}|)^\alpha + \\ &(|Y_{i,j-1}
     - Y_{i,j}| - |G(X)_{i,j-1} - G(X)_{i,j}|)^\alpha)
    \end{split}
    \end{equation}
    where $i$ and $j$ are the pixels' coordinates.
\end{itemize}

\subsection{Training}
In a first set of experiments, we used as discriminator the classifier described in Section~\ref{sec:classifier}, trained to distinghish between bad and good quality images, with fixed weights. In a second set of tests we allowed the weights of the discriminator to be  modified (as in the classical GAN approach), in order to further improve the obtained results.

We conducted six separate training sessions with the GAN model. In the first five, we progressively incorporated additional loss functions while keeping the discriminator fixed. 
The sixth session involved simultaneous training of both the generator and discriminator to compare its effectiveness against the fixed-discriminator configurations.

Each training was carried out using the ADAM optimizer with a learning rate of 0.001. Most sessions ran for 55 epochs, except the first (34 epochs) and the final joint-training experiment (20 epochs, due to increased computational cost).

For the first experiment, we used only the unpaired subset of the EUVP dataset, which includes 3,720 poor-quality images. For all subsequent sessions, we used 8,859 paired images (17,718 total), composed of:
\begin{itemize}
\item 700 paired samples from UIEB,
\item All available images from the Underwater Dark and Underwater Scenes subsets of EUVP,
\item 424 paired samples randomly selected from EUVP’s Underwater ImageNet subset (the remainder was reserved for later evaluation).
\end{itemize}

To increase robustness and dataset diversity, we applied data augmentation (horizontal and vertical flipping), resulting in a final training set of 53,154 images (26,577 per class).

The overarching objective of these sessions was to generate enhanced underwater images using real inputs, evaluated by a discriminator trained to reflect human-perceived quality. This introduces subjectivity into the adversarial learning process.

\textbf{Training with unpaired images and self-similarity loss (L1).}
In our first experiment, we trained the generator using only unpaired poor-quality images. To guide the generator without a reference target, we introduced an L1 loss that penalized deviation from the input image itself—encouraging preservation of structure and content:

\begin{equation} \label{funcionAdj}
Loss = L_{GAN}(G,D) + L1(X,X)
\end{equation}

\begin{figure}[h]
\begin{center}
	\includegraphics[width = 0.8\linewidth]{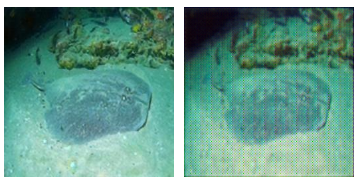}
\end{center}
\caption{Example of result obtained after training with loss function (\ref{funcionAdj}). Left, input. Right, generator output.}
\label{fig:resultadosAdv}
\end{figure}

As shown in Figure~\ref{fig:resultadosAdv}, although the discriminator assigned favorable scores (i.e., close to 0, indicating high quality), visual inspection of the generated images revealed poor results. This confirms a known weakness of adversarial models: discriminators can be deceived without true perceptual improvement in outputs.

\textbf{Training with paired images and L1 loss.}
Given the shortcomings of the previous approach, we switched to paired data and modified the loss to compare the generator’s output directly with the corresponding high-quality reference:

\begin{equation} \label{funcionL1Obj}
Loss = L_{GAN}(G,D) + L1(X,Y)
\end{equation}

\begin{figure}
\begin{center}
	\includegraphics[width = \linewidth]{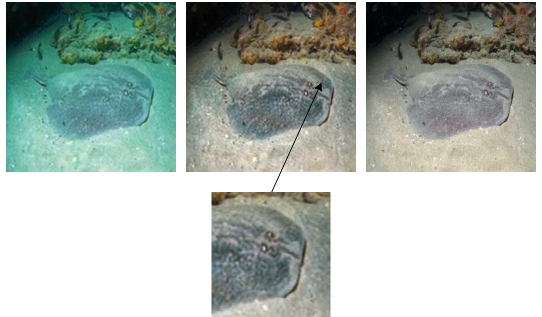}
\end{center}
\caption{Example of result obtained after training with loss function (\ref{funcionL1Obj}). Left, input. Center, generator output (with a detail shown below). Right, target.}

\label{fig:resultadosL1}
\end{figure}

This yielded significantly improved results (Figure~\ref{fig:resultadosL1}). Classification scores dropped to ~0.2, and the generator output began to visually resemble the target images. We selected the model from epoch 20, where convergence was observed and overfitting avoided. Minor artifacts and blurring remained, motivating further loss function enhancements.

\textbf{Training with paired images and L2 loss.}
To investigate the effect of using L2 instead of L1 loss, we modified the objective as follows:

\begin{equation} \label{funcionL2Obj}
Loss = L_{GAN}(G,D) + L2(X,Y)
\end{equation}

\begin{figure}
\begin{center}
	\includegraphics[width = \linewidth]{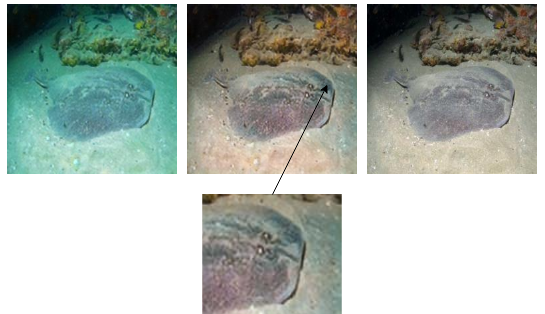}
\end{center}
\caption{Example of result obtained after training with loss function (\ref{funcionL2Obj}). Left, input. Center, generator output (with a detail shown below). Right, target.}

\label{fig:resultadosL2}
\end{figure}

As seen in Figure~\ref{fig:resultadosL2}, results were comparable to the L1 experiment. While some smoothing was observed, minor inconsistencies in color and sharpness remained.

\textbf{Training with L2 and Angular loss (L2A)}

To improve color fidelity, we extended the loss function to include angular loss, promoting natural RGB color alignment:

\begin{equation} \label{funcionL2Obj+}
Loss = L_{GAN}(G,D) + L2(X,Y) + \lambda_{\text{ang}} L_{\text{angular}}(X,Y)
\end{equation}
where $\lambda_{\text{ang}} = 0.8$.

\begin{figure}[h]
\begin{center}
	\includegraphics[width = 170px]{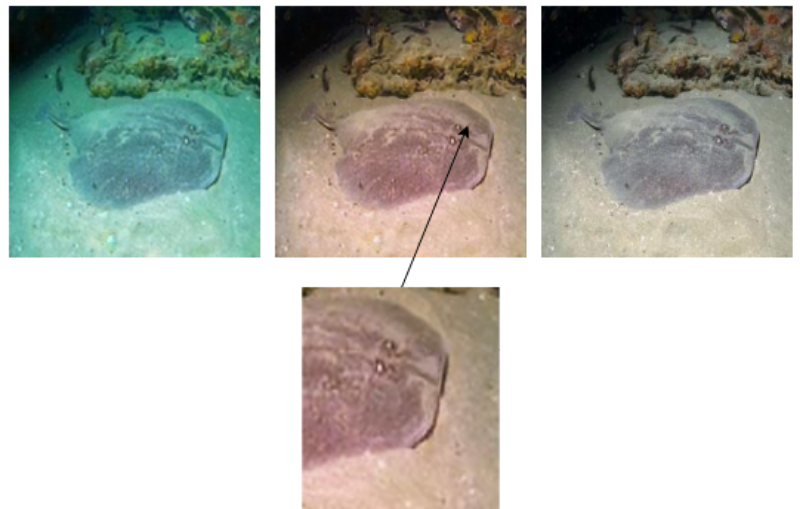}
\end{center}
\caption{Example of result obtained after training with loss function (\ref{funcionL2Obj+}). Left, input. Center, generator output (with a detail shown below). Right, target.}
\label{fig:resultadosL2A}
\end{figure}

This modification led to modest color improvements, as seen in Figure~\ref{fig:resultadosL2A}, with more consistent tone and reduced hue discontinuities.

\textbf{Training with L2, Angular, and Gradient loss (L2AG)}

To address image sharpness, we added gradient difference loss (GDL) to the objective function:

\begin{equation}
\label{funcionL2++}
\begin{array}{rl}
Loss = & L_{GAN}(G,D) + L2(X,Y) + \\ 
& +\lambda_{\text{ang}} L_{\text{angular}}(X,Y) + \\
& +\lambda_{\text{gdl}} L_{\text{gdl}}(X,Y)
\end{array}
\end{equation}

where $\lambda_{\text{ang}} = 0.8$ and $\lambda_{\text{gdl}} = 0.4$.

\begin{figure}[h]
\begin{center}
	\includegraphics[width = 170px]{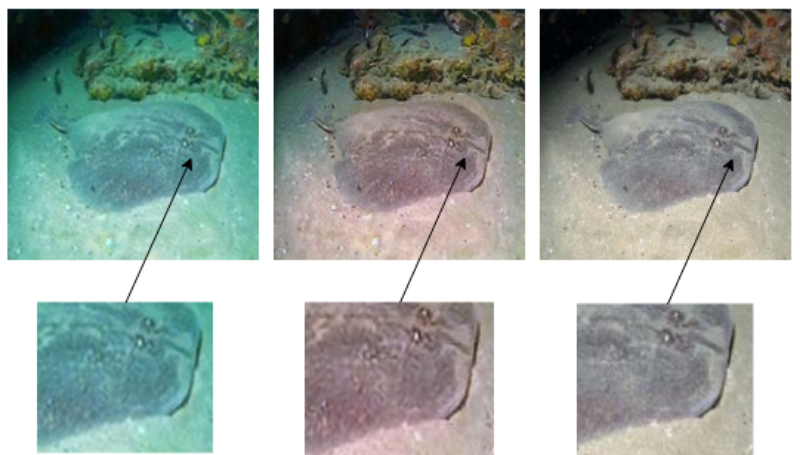}
\end{center}
\caption{Example of result obtained after training with loss function (\ref{funcionL2++}). Top row: Left, input. Center, generator output. Right, target. Bottom row: details}
\label{fig:resultadosL2AG}
\end{figure}

As shown in Figure~\ref{fig:resultadosL2AG}, image sharpness improved noticeably. Edges appeared clearer and artifacts reduced, with classification scores remaining near 0.2.

\textbf{Joint training of generator and discriminator (L2AGR)}

Finally, we retrained the discriminator alongside the generator using the same loss function (\ref{funcionL2++}). This was motivated by early observations that the fixed discriminator was easily deceived, consistently assigning low scores despite poor outputs.

\begin{figure}
\begin{center}
	\includegraphics[width = 180px]{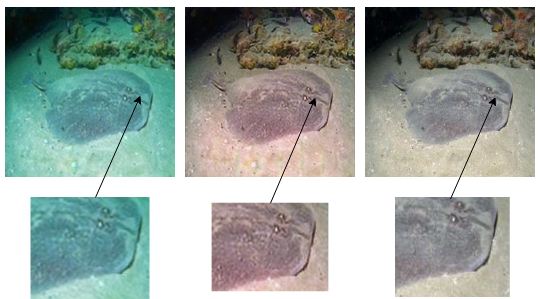}
\end{center}
\caption{Results from joint training with full loss function and dynamic discriminator. The loss function is (\ref{funcionL2++}). Top row: Left, input. Center, generator output. Right, target. Bottom row: details.}
\label{fig:resultadosL2AGR}
\end{figure}

As shown in Figure~\ref{fig:resultadosL2AGR}, this strategy led to greater variation in discriminator scores, reflecting its improved sensitivity. Visually, outputs exhibited enhanced sharpness and color consistency relative to earlier methods.

In the following section, we present a detailed evaluation and comparison of these training strategies using quantitative and qualitative metrics.

\section{Experiments}
\label{sec:tests}

All training and evaluation experiments were conducted on an Nvidia GeForce GTX 1060 GPU with 6 GB of RAM. The experiments are structured as follows: a quantitative evaluation using standard image quality metrics, a qualitative assessment, and a discussion of limitations.

\subsection{Quantitative Evaluation}

We evaluated the performance of our trained models using two widely adopted metrics in image reconstruction tasks: Peak Signal-to-Noise Ratio (PSNR) and Structural Similarity Index (SSIM)~\cite{metricasRuido}. These metrics compare the generated enhanced images with their corresponding high-quality reference images from a test set not used during training.

The PSNR metric is defined as:

\begin{equation}
    PSNR(x, y) = 10 \log_{10} \left( \frac{255^2}{MSE(x, y)} \right)
    \label{psnr}
\end{equation}

where $x$ is the generated image, $y$ is the reference image, and $MSE$ is the mean squared error between them. Higher PSNR values indicate better similarity to the reference image.

The SSIM metric~\cite{ruidoSSim} evaluates local structural similarity by comparing image patches based on luminance, contrast, and structural information:

\begin{equation}
    SSIM(x, y) = \frac{(2\mu_x \mu_y + c_1)(2\sigma_{xy} + c_2)}{(\mu_x^2 + \mu_y^2 + c_1)(\sigma_x^2 + \sigma_y^2 + c_2)}
    \label{ssim}
\end{equation}

In Equation~\ref{ssim}, $\mu_x$ and $\mu_y$ are the means of the image patches, $\sigma_x^2$ and $\sigma_y^2$ are their variances, and $\sigma_{xy}$ is the covariance. The constants $c_1 = (255 \times 0.01)^2$ and $c_2 = (255 \times 0.03)^2$ ensure numerical stability~\cite{EUVP}.

Table~\ref{tab:PSNR Y SSIM} presents the average PSNR and SSIM values over 984 test images for each model variant. The results show that the final training configuration (L2AGR) yields the best performance on both metrics.

\begin{table}[!htb]
    \centering
    \caption{Quantitative Comparison Using PSNR and SSIM on 984 Test Images.}
    \begin{tabular}{|c|c|c|}
        \hline
        Method & PSNR (dB) & SSIM \\
        \hline
        Input     & 24.93 & 0.76 \\
        L1        & 25.75 & 0.79 \\
        L2        & 26.09 & 0.79 \\
        L2A       & 26.13 & 0.79 \\
        L2AG      & 26.26 & 0.80 \\
        L2AGR     & \textbf{26.34} & \textbf{0.81} \\
        \hline
    \end{tabular}
    \label{tab:PSNR Y SSIM}
\end{table}

In addition to PSNR and SSIM, we assessed image quality using the Underwater Image Quality Measure (UIQM)~\cite{UCIQM}, which evaluates aspects specific to underwater imagery, including colorfulness, sharpness, and contrast. Higher values indicate better perceptual quality. Results are shown in Table~\ref{tab:UIQM}.

\begin{table}[!htb]
    \centering
    \caption{Quantitative Comparison Using UIQM on 984 Test Images.}
    \begin{tabular}{|c|c|}
        \hline
        Method & UIQM \\
        \hline
        Input     & 2.69 \\
        Goal      & 3.09 \\
        L1        & 2.77 \\
        L2        & 2.87 \\
        L2A       & 2.91 \\
        L2AG      & 2.92 \\
        L2AGR     & \textbf{3.05} \\
        \hline
    \end{tabular}
    \label{tab:UIQM}
\end{table}

These results highlight the improvement achieved through successive refinement of the loss functions. Specifically:
\begin{itemize}
\item PSNR and SSIM: L2AGR achieves a 2\% improvement over L1 and a 5–6\% improvement over the original input images.
\item UIQM: L2AGR shows a 10\% improvement over L1 and a 13\% improvement over the input, reaching within 1.3\% of the reference (goal) images—suggesting a perceptual quality close to that of the high-quality targets.
\end{itemize}

\subsection{Qualitative Evaluation}

\begin{figure*}
\begin{center}
	\includegraphics[width=\linewidth]{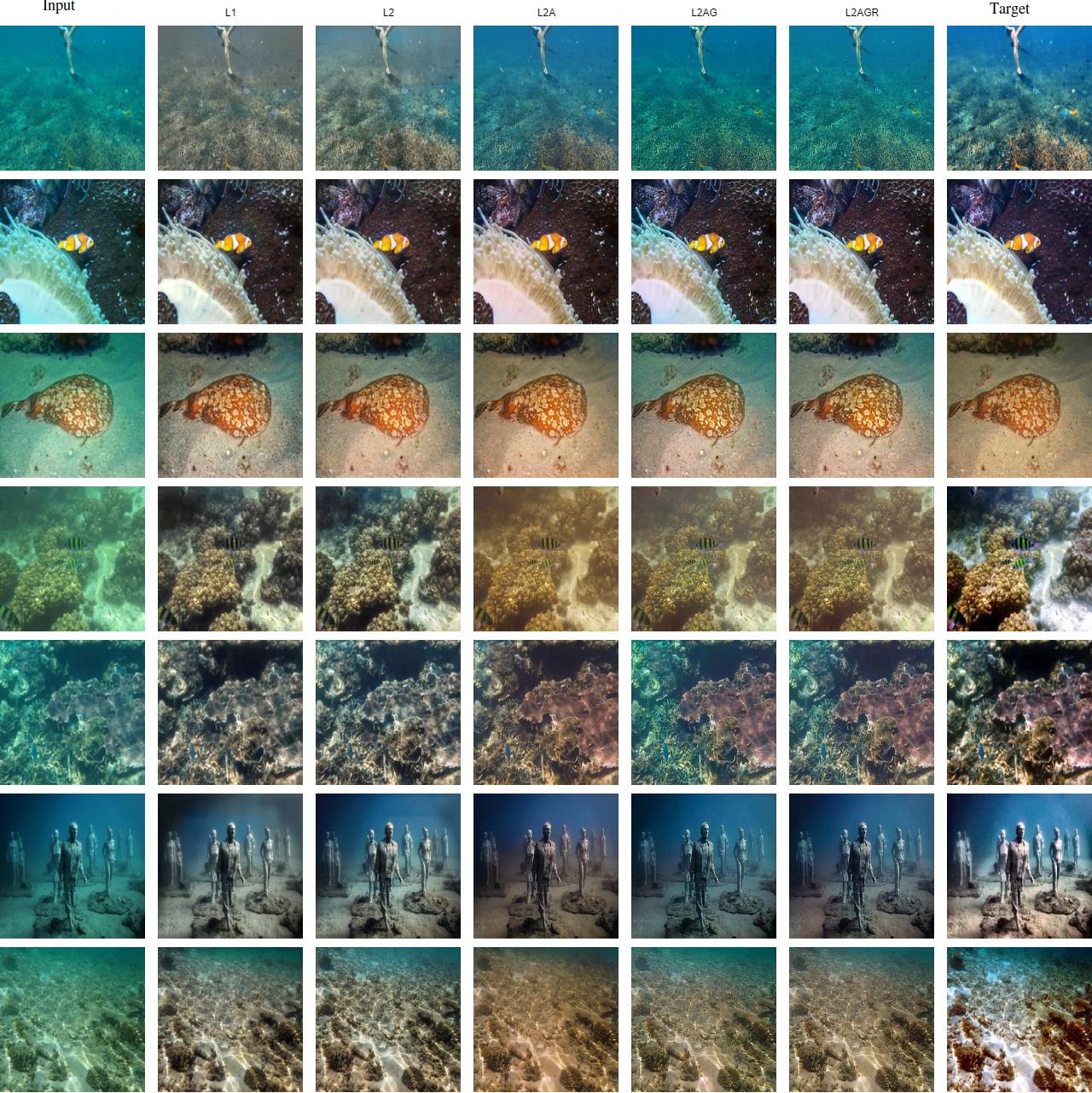}
\end{center}
\caption{Visual comparison of results from different training configurations alongside the input and reference images.}
\label{ImagenesVisual}
\end{figure*}

Given that our objective involves enhancing underwater images using a generative network, quantitative evaluation alone is insufficient. Human visual perception plays a fundamental role in determining the success of the enhancement, as image quality should also be judged based on realism, aesthetic appeal, and visual coherence.

To this end, we conducted a qualitative assessment by visually comparing the outputs of each trained generator against the original input and target reference images. Figure~\ref{ImagenesVisual} presents a selection of input images processed by each model, arranged for direct visual comparison with their corresponding ground truths.

From this visual analysis, it is evident that all models produced enhancements that are perceptually superior to the input images. Specifically, improvements were observed in color correction and, to a lesser extent, in sharpness. Across all models, the enhanced outputs are more visually aligned with the reference images in both tone and contrast.

Among the various configurations, the L2AGR model yielded the most visually compelling results, with images that not only approximate the target in color fidelity but also exhibit improved sharpness and fewer artifacts. These findings, combined with the quantitative metrics reported earlier, confirm the effectiveness of the proposed training strategy in improving the perceptual quality of underwater images.

\subsection{Limitations}

\begin{figure*}
\begin{center}
	\includegraphics[width=\linewidth]{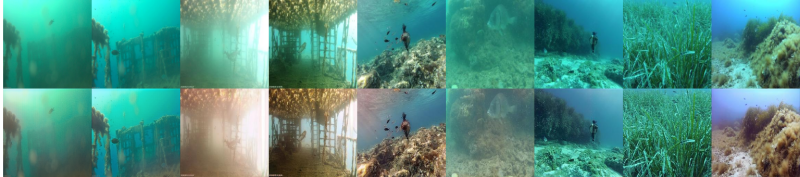}
\end{center}
\caption{Example outputs generated by the L2AGR model on highly turbid underwater images.Top row: input. Bottom row: enhancement result.}
\label{ResultFInal}
\end{figure*}

To further evaluate the generalizability of the proposed model, we conducted an additional experiment using real underwater images exhibiting high turbidity (haze). This test revealed an important limitation of the L2AGR model, which was otherwise the most effective configuration in our study. As shown in Figure~\ref{ResultFInal}, although the model achieves partial enhancement in some cases, it fails to consistently produce high-quality results under extreme haze conditions.

This limitation can be attributed to a lack of high-turbidity samples in the training data. Since the model was not exposed to such heavily degraded images during training, its ability to generalize to these conditions is constrained. The presence of haze introduces complex distortions—such as reduced contrast, color shifting, and loss of fine details—that the model was not explicitly trained to resolve.

Nevertheless, it is worth noting that the L2AGR model still manages to improve visual quality to some degree, even in these difficult cases. This suggests that the proposed training strategy retains some robustness to unseen distortions, though further improvements are needed.

Future work may address this limitation by incorporating targeted data augmentation, synthetic haze modeling, or domain adaptation techniques to enhance the model’s ability to process more severely degraded underwater imagery.

\section{Conclusions}
\label{sec:conclusions}

This study addressed two key challenges in underwater image enhancement. The first was the development of a deep learning-based classifier capable of distinguishing between high- and low-quality marine images. This model serves a dual purpose: (1) supporting the enhancement pipeline by acting as a quality assessor, and (2) enabling the automatic generation of labeled datasets for future use. Notably, the classifier was trained solely on real images labeled by expert consensus, incorporating human subjectivity into the learning process.

The second challenge was the improvement of underwater image quality through deep learning techniques. We explored adversarial training via GANs, with the goal of producing high-quality outputs without relying on artificially generated in-air references.

The results demonstrate clear progress in enhancing the visual and perceptual quality of underwater images. Both quantitative metrics and qualitative evaluation confirm that all training strategies contributed to improvements over the input images. Among them, the final model—L2AGR—achieved the best results, showing consistent gains in PSNR, SSIM, and UIQM, as well as in visual realism based on human judgment.

A key limitation observed was the model's reduced effectiveness in handling images with high turbidity. This issue is attributed to a lack of such examples in the training dataset. These findings point to the importance of incorporating a more diverse range of underwater conditions into training data and developing models that are more robust to challenging distortions such as haze.

Overall, the study demonstrates the viability of using deep learning techniques—particularly GAN-based architectures informed by subjective image quality—to enhance real underwater imagery. With potential applications in marine research, underwater exploration, and conservation, continued development in this area could contribute meaningfully to efforts aimed at understanding and preserving marine ecosystems.

\section*{Acknowledgment}
The publication is part of the project PID2021-125711OB-I00, financed by MCIN/AEI/10.13039/501100011033/FEDER, EU.
This work has been partially sponsored and promoted by the Comunitat Autonoma de les Illes Balears through the Conselleria d’Educació I Universitats and by the European Union- Next Generation EU (BIO2022/002A.1, 022B.1) and by Plans complementaris del Pla de Recuperaci\'o, Transformaci\'o i Resili\`encia (PRTR-C17-I1). Nevertheless, the views and opinions expressed are solely those of the author or authors, and do not necessarily reflect those of the Conselleria d’Educació i Universitats, the European Union or the European Commission. Therefore, none of these organisations are to be held responsible.

\bibliographystyle{siam}
\bibliography{article}

\end{document}